\documentclass[letterpaper, 10 pt, conference]{ieeeconf}  % Comment this line out if you need a4paper

\IEEEoverridecommandlockouts                              % This command is only needed if 
\usepackage{epsfig} % for postscript graphics files
\usepackage{times} % assumes new font selection scheme installed
\usepackage{amsmath} % assumes amsmath package installed
\usepackage{amssymb}  % assumes amsmath package installed
\usepackage{color}
\usepackage{xcolor}
\usepackage{graphicx}
\usepackage{url} 
\usepackage[format=plain,labelsep=period]{caption}
\usepackage{wrapfig}
\usepackage[list=on,listformat=simple]{subcaption}
\usepackage{makecell}
\usepackage{float}
\floatstyle{plaintop}
\restylefloat{table}
\usepackage{mathtools}
\usepackage{hyperref}
\usepackage{cite}

\usepackage{preambles}

\title{\LARGE \bf
Reinforcement Learning with Data Bootstrapping for Dynamic Subgoal Pursuit in Humanoid Robot Navigation}
\author{Chengyang Peng$^{1}$, Zhihao Zhang$^{2}$, Shiting Gong$^{3}$, Sankalp Agrawal$^{2}$, Keith A. Redmill$^{2}$, and Ayonga Hereid$^{1}$% <-this % stops a space
\thanks{*This work was supported in part by the National Science Foundation under grant FRR-21441568. }% 
\thanks{$^{1}$Mechanical and Aerospace Engineering, The Ohio State University, Columbus, OH, USA. {\tt\footnotesize (peng.947, hereid.1)@osu.edu.}}%
\thanks{$^{2}$Electrical and Computer Engineering, The Ohio State University, Columbus, OH, USA. {\tt\footnotesize (zhang.11606, agrawal.268, redmill.1)@osu.edu.}}%
\thanks{$^{3}$Computer Science and Engineering, The Ohio State University, Columbus, OH, USA. {\tt\footnotesize (gong.663)@osu.edu.}}%
}

\begin{document}

\maketitle
\thispagestyle{empty}
\pagestyle{empty}

\begin{abstract}
Safe and real-time navigation is fundamental for humanoid robot applications. However, existing bipedal robot navigation frameworks often struggle to balance computational efficiency with the precision required for stable locomotion. We propose a novel hierarchical framework that continuously generates dynamic subgoals to guide the robot through cluttered environments. Our method comprises a high-level reinforcement learning (RL) planner for subgoal selection in a robot-centric coordinate system and a low-level Model Predictive Control (MPC) based planner which produces robust walking gaits to reach these subgoals. To expedite and stabilize the training process, we incorporate a data bootstrapping technique that leverages a model-based navigation approach to generate a diverse, informative dataset. We validate our method in simulation using the Agility Robotics Digit humanoid across multiple scenarios with random obstacles. Results show that our framework significantly improves navigation success rates and adaptability compared to both the original model-based method and other learning-based methods.

% Safe path planning and following is critial for bipedal robot navigation. However, in traditional navigation problem, path planning and path following are often treated independently, potentially causing unsafe situations during the robot's movement. Moreover, a safety-critical path must consider not only avoiding obstacles but also ensuring the bipedal robot's walking adheres to its kinematic constraints.
% This work introduces an Enhanced Safety-Critical (ESC) planner that achieves the unification of path planning and control planning. The planner integrates the Linear Inverted Pendulum (LIP)-based Model Predictive Control (MPC), considering both environmental constraints and robot kinematics constraints. To validate its ability for bipedal robot controlling, we also introduced a hierarchical controller for bipedal safe navigation that integrates the LIP model-based ESC planner as a high-level planner for generating task space commands and a low-level controller for task space tracking. In the simulation, our hierarchical framework successfully controlled the robot to achieve efficient navigation and safe avoidance in various obstacle environments, demonstrating improved performance compared to the same structure controller but with a differential drive model.
\end{abstract}

\section{Introduction}

Humanoid robots are increasingly demonstrating their adaptability to complex and crowded environments. Equipped with dexterous manipulation and versatile locomotion capabilities, these robots can navigate human environments and handle objects with precision, making them well-suited for tasks in logistics, public services, and entertainment. Because these tasks frequently take place in crowded environments, robust navigation capabilities are essential to ensure safe, efficient, and reliable performance. However, navigating these dynamic and often unpredictable settings poses significant challenges for bipedal robots given their underactuated and high-dimensional dynamics~\cite{Wight2008Foot, wermelinger2016navigation, kuindersma2016optimization}.
Current navigation frameworks commonly separate geometric path planning from real-time locomotion control, making it difficult to accommodate the nonlinear dynamics of bipedal robots and limiting their performance in unpredictable or crowded environment~\cite{sleumer1999exact,sariff2006overview,gasparetto2015path,kim2019confidence}. Moreover, discrepancies between the planned path and the executed trajectory are inevitable due to model inaccuracies, external disturbances, and dynamic constraints. Such deviations are especially critical for underactuated bipedal robots, where even minor errors can lead to instability or falls, underscoring the need for robust navigation methods \cite{winkler2018optimization,shamsah2023integrated,peng2024unified}.

% Optimization-based methods integrate path planning with robot dynamics by formulating and solving constrained optimization problems, making them widely used in bipedal robot locomotion control. many studies have proposed trajectory optimization controllers based on whole-body dynamics [cite] or reduced order dynamics [cite], demonstrating significant improvements in motion performance. These frameworks primarily focus on the robot’s center of mass (CoM) controlling by giving specific velocity or turning tasks, and determine foot placement or contact forces through Model Predictive Control (MPC) or Quadratic Programming (QP)-based optimization.  However, the navigation problem requires these tasks to be determined based on the environment, which introduces a critical challenge that the coupling of turning rate and heading angle with other states leads to nonlinearities of the problem and complicates the optimization process [cite]. Although, in our previous work, we introduced a novel way to linearize the optimization problem by precomputing the terms causing nonlinearities, the proposed framework still limits the navigation flexibility of bipedal robots.

\begin{figure}
\centering
\vspace{2mm}
    \includegraphics[trim={0cm 0cm 0cm 0cm},clip,width=1\columnwidth]{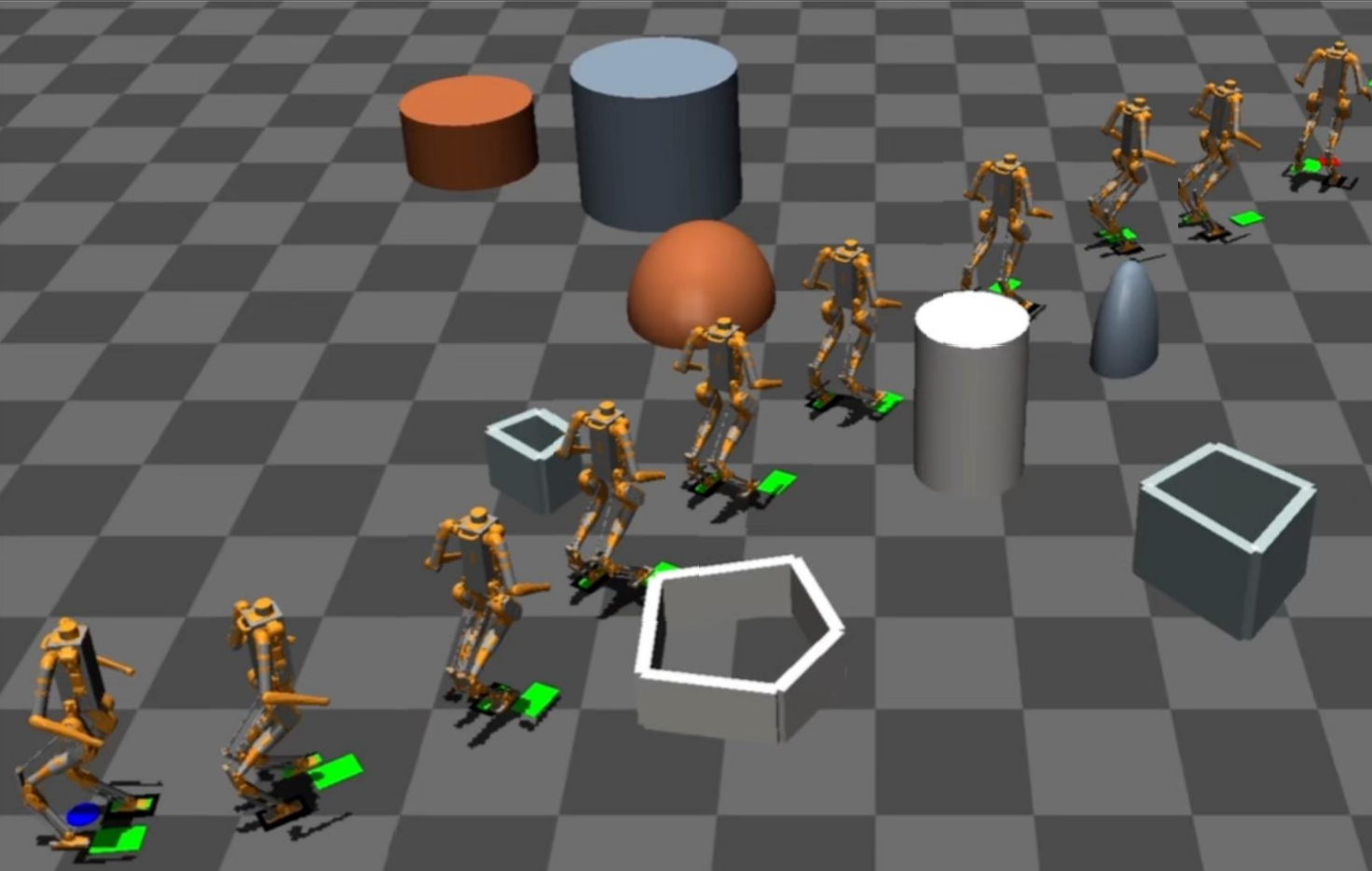}
\caption{\small{A humanoid robot navigates around clustered environment using our proposed sub-goal orientated framework.}
} 
\label{fig:simu_nav_1}
\vspace{-3mm}
\end{figure}

Model-based methods are commonly used in bipedal locomotion control, formulating and solving constrained optimization problems to integrate path planning with robot dynamics. Many studies have developed trajectory optimization controllers based on whole-body dynamics  \cite{diehl2006fast,mombaur2009using,dai2014whole} or reduced-order dynamics \cite{kajita20013d, englsberger2011bipedal,rutschmann2012nonlinear,grandia2023perceptive}, leading to significant improvements in motion performance. In practice, these frameworks rely on simplified center-of-mass or foot-placement models, using Model Predictive Control (MPC) or Quadratic Programming (QP) to compute stable walking gaits. However, effectively integrating high-dimensional environmental information (e.g., occupancy maps) into these optimization problems is challenging, as it greatly increases the state space and number of constraints—ultimately slowing real-time computation. Moreover, the nonlinear coupling of the heading angle, turning rate, and other system states further complicates the optimization process~\cite{griffin2016model, narkhede2022sequential, peng2024unified}. While our previous work \cite{peng2024real} attempted to mitigate these nonlinearities by precomputing heading angles, it still relies on a geometric approach and thus cannot fully leverage the flexibility of bipedal locomotion in cluttered environments.

\begin{figure*}
\centering
\vspace{2mm}
    \includegraphics[trim={0cm 0cm 0.2cm 0.5cm},clip,width=0.96\textwidth]{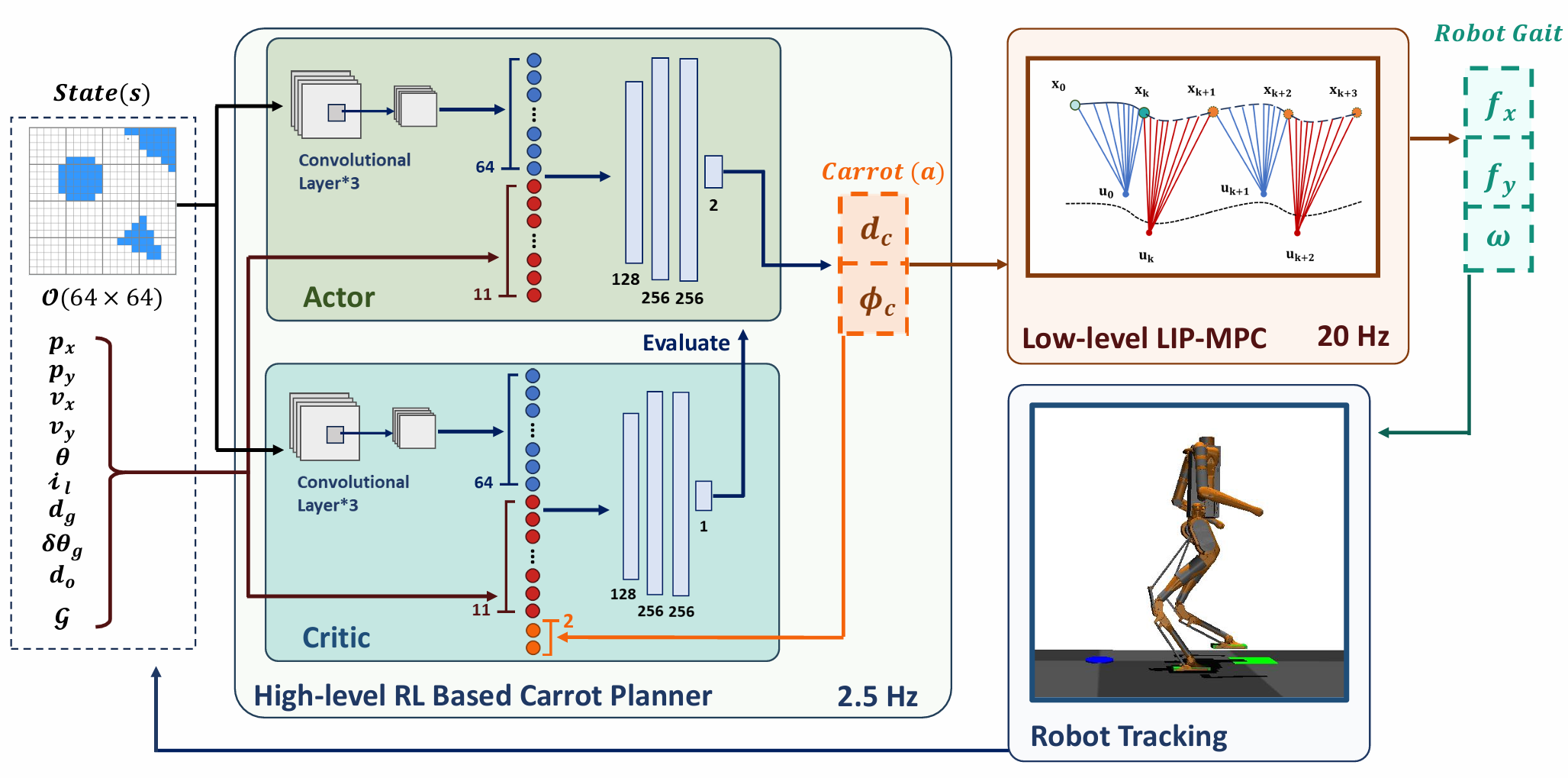}
\caption{Overall structure of the proposed hierarchical framework for humanoid navigation. The high-level RL-based planner uses local map data to continuously generate feasible dynamic subgoals that guide the robot step-by-step toward the global target, while the low-level MPC-based controller computes dynamically stable walking gaits that satisfy both motion and obstacle constraints to follow these subgoals.}
\label{fig:structure}
\vspace{-6mm}
\end{figure*}

Recent advances in learning-based methods have showcased their potential in handling nonlinearities and optimizing high-dimensional control problems, opening new avenues for addressing complex robot navigation challenges. Imitation learning, for instance, leverages deep neural networks to train policies that mimic expert behavior \cite{abbeel2004apprenticeship,syed2008apprenticeship,ho2016generative}. However, this approach relies heavily on high-quality expert data and lacks active environmental interaction, limiting its ability to generalize to novel scenarios. Reinforcement learning (RL) offers an alternative by enabling robots to learn navigation strategies through trial and error. RL-based approaches have shown promise in adaptive decision-making and motion planning \cite{cimurs2020goal,brito2021go,perez2021robot,zhu2022hierarchical,xie2023drl}. Despite these advances, most existing RL navigation frameworks are tailored for wheeled robots with simpler kinematics and more straightforward action spaces. Moreover, RL methods are often hindered by sample inefficiency, as training from scratch without leveraging prior knowledge of the environment and robotic dynamics is challenging \cite{huang2022efficient,yuan2024pre}. This issue becomes even more pronounced in continuous action spaces and complex environments, where sparse or delayed reward signals increase the likelihood of converging to suboptimal solutions.

% The second strategy, supported by recent studies \cite{hester2018deep}, involves incorporating demonstration data into the replay buffer during training. In our implementation, the replay buffer is populated with samples drawn from both online experience and the offline demonstration dataset, thereby effectively enhancing the RL training process.

To bridge the gap between RL-based navigation and bipedal robot locomotion, we propose a novel hierarchical RL+model-based framework that divides the navigation task into two levels. At the high level, an RL planner dynamically generates subgoals in a local, robot-centric coordinate frame. At the low level, a gait planner based on a linear MPC leverages reduced-order bipedal dynamics to maintain dynamic stability and produce robust walking gaits for following these subgoals. By combining the adaptability of RL, capable of handling complex environments and unforeseen disturbances, with MPC’s predictive control of future states, our framework helps mitigate the risk of instability or falls in challenging scenarios. Furthermore, we adopt a demonstration data bootstrapping strategy, consistent with recent approaches in the literature \cite{hester2018deep}. Specifically, our replay buffer is populated with both online experiences and samples from an offline demonstration dataset generated by a model-based navigation method, which accelerates learning and enhances the robustness of the learned navigation policies.
% To mitigate computational efficiency and navigation flexibility, the subgoal direction, one output of the RL planner, is used to precompute the turning rate for each prediction step, allowing us to linearize the constraints within the low-level MPC. 
% Since RL updates its decisions step by step, the navigation process resembles chasing a dangling carrot, inspiring us to name our approach “Carrot Chasing.”

The rest of the paper is organized as follows: Section~\ref{sec:method} details our Subgoal-Pursuit hierarchical framework. Section~\ref{sec:setup} introduces the training setup and procedure of the framework.
Section~\ref{sec:sim} evaluates its performance in simulation, comparing it with model-based frameworks and various state-of-the-art learning-based high-level planners. Finally, Section~\ref{sec:conclusion} concludes the contributions, limitations, and future work.

\section{Hierarchical Bipedal Navigation Framework}
\label{sec:method}

This section introduces a two-level hierarchical architecture for robust bipedal navigation and obstacle avoidance, as illustrated in \figref{fig:structure}. The central idea is to learn dynamic subgoals using only local map data and other state information, enabling the robot to navigate complex environments.

% we present the methodology of our proposed dynamic subgoal pursuit framework for bipedal navigation and effective obstacle avoidance. As illustrated in \figref{fig:structure}, our framework adopts a 

\subsection{Reinforcement Learning for Dynamic Subgoal Planner}
We formulate the problem of learning dynamic subgoals for bipedal navigation as a Markov Decision Process (MDP) defined by the tuple:
\begin{align}
\label{eq:mdp}
    \mathcal{M} = (\mathcal{S}, \mathcal{A}, \mathcal{P}, r, \gamma),
\end{align}
where $\mathcal{S}$ is the state space; $\mathcal{A}$ is the action space; $\mathcal{P}$ denotes the state transition model; $r :\mathcal{S}\times\mathcal{A}\xrightarrow{}\mathbb{R}$ is a reward function and $\gamma \in (0,1]$ is a discount factor.

\newsec{State Space.}
To generate effective dynamic subgoals for navigation, the high-level RL planner must capture both environmental perception and the robot’s dynamic state. In this work, we assume that the robot’s perception is limited to a relatively small local region.
% , which it relies on for effective obstacle avoidance.
To facilitate this, we construct a 64×64 local occupancy grid map centered on the robot, which spans a forward detection range of 4.5 m, a backward detection range of 1.5 m, and a lateral width of 6 m. This occupancy map, denoted as $\mathcal{O}$, provides structured local perception data that enables the RL agent to accurately detect obstacles and plan safe maneuvers. In addition to the environmental data, the robot’s locomotion state is a key feedback component of the state representation.
Since the high-level planner generates a new subgoal at the beginning of each walking step, we represent the robot’s reduced-dimensional step-to-step dynamics by capturing its center-of-mass (CoM) state at the start of each step. 
Moreover, to steer the robot toward a designated goal, we incorporate goal-related information—specifically, the goal position and the corresponding Euclidean distance and relative heading to the goal—into the state. Thus, the final state $s\in \mathcal{S}$ is defined as follows:
\begin{align}
\label{eq:rl_state}
    s = \{\mathcal{O}, p_x, p_y, v_x, v_y, \theta,i_{l}, d_{g}, \delta\theta_{g}, d_{o}, \mathcal{G}\},
\end{align}
where $\mathcal{O}$ denotes the occupancy map, $[p_x, p_y, v_x, v_y, \theta]$ represent the robot’s CoM position, velocity, and heading angle in the world frame. $i_{l}$ indicates the stance foot, with $-1$ for the left foot and $1$ for the right foot. $[d_{g}, \delta\theta_{g}]$ denote the robot’s Euclidean distance and relative heading to the goal, while $d_{o}$ measures the distance to the nearest obstacle in the local map. Finally, $\mathcal{G}$ represents the goal position in the world frame.

\newsec{Action Space.} The high-level planner defines its action $a\in\mathcal{A}$ as a dynamic subgoal specified in a robot-centric polar coordinate system. The action is represented as:
\begin{align} 
\label{eq:rl_act}
a = (d_c, \phi_c), 
\end{align}
where $d_c$ denotes the distance from the robot to the subgoal and $\phi_c$ indicates the direction relative to the robot’s current heading. These parameters are constrained by the robot’s physical capabilities, such as step length and turning limits, to ensure compatibility with the low-level MPC controller. In our implementation, $d_c$ is limited to $[0, 3]$ meters, and $\phi_c$ is limited to $[-\pi/4, \pi/4]$ radians.

To model the stochastic policy $\pi(a|s)$ in the continuous action space, we use the Soft Actor-Critic (SAC) \cite{haarnoja2018soft} algorithm. Specifically, the policy $\pi$ is modeled as a Gaussian distribution, where actions are sampled as:
\begin{align} 
\label{eq:rl_act_norm}
a \sim \pi(a|s) = \mathcal{N}(\mu_\pi(s), \sigma_\pi(s)). 
\end{align}
Here, $\mu_\pi(s)$ and $\sigma_\pi(s)$ are the mean and standard deviation of the action distribution, which are the output of the policy network. The optimal policy is obtained by maximizing the expected cumulative discounted reward, expressed as:
\begin{align*} 
% \label{eq:rl_policy}
\pi^* = \arg \max_\pi  \mathbb{E}_\pi \left[\sum_t \gamma^t r(s[t], a[t]) +\alpha H(\pi(\cdot|s[t]))\right],
\end{align*}
where $H(\pi(\cdot|s_t))$ represents the entropy term of the policy and $\alpha$ is the regularization factor. The policy maintains a balance between long-term optimality and sufficient exploration by leveraging SAC's entropy-regularized objective, ensuring robust navigation performance.

\newsec{Network Structure.} 
Since we use the SAC algorithm, our framework utilizes two neural networks: an actor network and a critic network. The actor network takes as input the heterogeneous state representation defined in \eqref{eq:rl_state} and generates dynamic subgoals---i.e., actions---according to \eqref{eq:rl_act_norm}. To process the heterogeneous data, the actor network is composed of two modules. First, a Convolutional Neural Network (CNN) processes the occupancy map. This CNN has three convolutional layers with $3\times3$ kernels, each followed by ReLU activation and max pooling to extract spatial features. The output of the final convolutional layer is flattened and passed through two fully connected layers, reducing the feature dimension to 64. Next, these 64-dimensional map features are concatenated with the remaining state information and are processed by four fully connected layers which output the parameters of a Gaussian distribution over the dynamic subgoals (actions). The critic network adopts a similar architecture, with the only difference being that its final output is a single scalar value representing the state-action value. The structure of the proposed actor-critic network is illustrated in \figref{fig:structure}.

\newsec{Reward Function.} Designing an effective reward function is crucial for reinforcement learning-based navigation, as it directly influences the generation of feasible dynamic subgoals that guide the robot toward the goal while ensuring obstacle avoidance and smooth locomotion. To achieve this, we define the total reward function as a weighted sum of several sub-rewards:
\begin{align} 
\label{eq:reward}
r = \mathbf{w} [r_{g}, r_{\theta}, r_{a}, r_{v}, r_o]^T+ R_{F},
\end{align}
where $r_g,r_{\theta},r_a,r_v$ and $r_o$ are sub-reward functions that address key aspects of bipedal navigation. Each reward term is normalized within $[0,1]$ to maintain numerical stability during training. In this work, we choose the weight vector as $\mathbf{w}=[0.2, 0.1, 0.25, 0.2, 0.25]$ to ensure a balanced learning process. $R_F$ is the terminal reward (and penalty) computed at the end of a navigation episode.

\newitem{Goal Proximity Reward}: To encourage goal-directed movement, we introduce a goal proximity reward $r_{g}$ that measures the reduction in distance $\Delta d_{g} = d_g[t-1]-d_g[t]$ to the goal between consecutive time steps. This reward is defined as:
\begin{align} 
\label{eq:reward_goal}
r_{g}[t] = \begin{cases}
      b_g+a_g\Delta d_{g}[t] & \text{if $\Delta d_{g}[t]\geq0$}\\
      0 &  \text{otherwise}
    \end{cases},
\end{align}
% \begin{align} 
% \label{eq:reward_goal}
% r_{g}[t] = \begin{cases}
%       b_g+a_g(d_g[t-1]-d_g[t]) & \text{if $d_g[t-1]-d_g[t]\geq0$}\\
%       0 &  \text{otherwise}
%     \end{cases},
% \end{align}
where $b_g$ is a baseline reward granted when the distance to the goal does not increase and $a_g$ is a scaling factor that encourages a greater reduction in the goal distance.

\newitem{Heading Alignment Reward}: 
To promote proper alignment of the robot’s orientation with the goal direction, we introduce a heading alignment reward $r_{\theta}$. We use a cubic function to penalize large heading errors:
\begin{align} 
\label{eq:reward_heading}
r_{\theta}[t] = \begin{cases}
      1-a_{\theta}\left\|\delta \theta_{g}[t]\right\|^3 & \text{if $\left\|\delta \theta_{g}[t]\right\|\leq \pi/6$}\\
      0 &  \text{otherwise}
    \end{cases},
\end{align}
where $a_{\theta}$ is a scaling factor that penalizes deviations from the goal direction.

\newitem{Action Smoothness Reward}: To promote smooth motion, we define an action smoothness reward $r_{a}$ that encourages consistent turning and gradual velocity changes. This reward consists of two components: one that incentivizes forward motion and another that penalizes abrupt action variations. The reward is formulated as:
\begin{align} 
\label{eq:reward_action}
r_{a}[t] &= q_a r_p [t]+(1-q_a)r_s[t],
\end{align}
where $r_p:= \|d_{c}\|_d-\|\phi_{c}\|_{\phi}$ encourages the robot to make progress toward the goal while minimizing unnecessary turns, whereas $r_{s}[t]:=1-\left(a[t]-a[t-1]\right)$ penalizes large deviations between consecutive actions to ensure smooth transitions, and $q_a \in [0,1]$ is a scaling factor that balances the weight for each component. 
% \begin{align} 
% \label{eq:reward_action_2}
% r_{{a_1}_t} &= \|d_{c_t}\|_d-\|\phi_{c_t}\|_{\phi} \\
% r_{{a_2}_t} &= 1-(a_t-a_{t-1}).
% \end{align}
Since the distance $d_{c}$ and heading $\phi_{c}$ are defined on different scales, we use the separate normalization functions, $\|\cdot\|_d$ for distance and $\|\cdot\|_{\phi}$ for heading, to bring them to a same scale. 

\newitem{Velocity Reward}: Since bipedal locomotion involves both longitudinal ($v_x$) and lateral ($v_y$) velocity components, we introduce a velocity reward $r_v$ that encourages maintaining longitudinal velocity and discourages lateral movement. This reward is formulated as follows:
\begin{align} 
\label{eq:reward_vel}
r_{v}[t] &= q_v r_{v_x}[t]+(1-q_v) r_{v_y}[t],
\end{align}
where $q_v$ is a weighting factor that balances the contributions of the two velocity components, and
\begin{align} 
\label{eq:reward_vel_2}
r_{v_x}[t] &= \frac{1}{1+e^{-a_v(v_x[t]-b_v)}},\\
r_{v_y}[t] &= \begin{cases}
      1 & \text{if $\|v_y[t]\|\leq 0.4$}\\
      0 &  \text{otherwise}
    \end{cases},
\end{align}
where $a_v$ is a scaling factor that encourages faster forward velocities and $b_v$ is a baseline (recommended) forward velocity. This formulation ensures that the robot is incentivized to maintain a desired forward velocity while minimizing undesirable lateral movements.

\newitem{Collision Avoidance Reward}: To enhance safety, we incorporate a collision avoidance reward $r_o$, inspired by our previous work with control barrier functions~\cite{peng2024real}. This reward is formulated using the half-plane distance to the closest obstacle:
\begin{align} 
\label{eq:reward_cbf}
\hspace{-2mm}
r_{o}[t] &= h(p_{x}[t], p_{y}[t])+(\zeta-1)h(p_{x}[t-1], p_{y}[t-1]),
\end{align}
where $h(p_{x}[t], p_{y}[t])$ is the linear function that measures the distance to the half-plane of the closest obstacles at time $t$, and $\zeta \in (0,1]$ is a scaling parameter of the barrier condition. This formulation rewards the robot for increasing its distance from obstacles over consecutive time steps, thereby promoting safer navigation.

\newitem{Terminal Reward and Penalty.}
In bipedal robot navigation, several terminal conditions can occur, including reaching the goal, colliding with obstacles, falling, and time-out. To account for these scenarios, we design specific rewards and penalties as follows:
\begin{align} 
\label{eq:reward_terminal}
R_{F} &= \begin{cases}
      60e^{-\frac{0.4\cdot N_{step}}{T_{max}}}+40 & \text{Goal Reached}\\
      -80 &  \text{Failure}\\
      -70 &  \text{Time-out}\\
    \end{cases},
\end{align}
where $N_{step}$ stands for the number of steps taken before reaching the goal and $T_{max}$ is the maximum time length for each training episode. The goal-reaching reward incorporates an exponential decay based on the number of steps to encourage faster navigation, while a large penalty is assigned to failure cases, such as colliding and falling. 
A slightly lower penalty is assigned to time-out scenarios, as it suggests the agent continues to explore rather than failing abruptly.

% \subsubsection{Model-based Controller as Demonstration}
% In our previous work, we developed a model-based nonlinear MPC controller for bipedal robot navigation in complex, cluttered environments. Despite its effectiveness, the computational expense of this approach (further detailed in Section III.B) renders it impractical for providing online step decisions or guidance during RL training\cite{wu2022prioritized}. Instead, we utilize the nonlinear MPC controller to generate an offline demonstration dataset.

% Two principal strategies exist for bootstrapping RL training with demonstration data. The first strategy is to recover a control policy from the dataset through behavior cloning, potentially augmented with an entropy-based penalty in the reward function\cite{yuan2024pre}. However, as detailed in Section [insert section reference], the limited size of the dataset combined with the problem's inherent complexity prevented the recovery of a robust control policy via BC.

\subsection{LIP based linear MPC for Stable Gait Planning}
The low-level MPC-based gait controller receives a subgoal generated by the high-level RL planner at the beginning of each step and generates a sequence of walking gaits based on the reduced dimensional Linear Inverted Pendulum (LIP) model~\cite{kajita20013d, peng2024unified, peng2024real}. 

Under the assumption that the robot’s CoM height and centroidal momentum remain constant, the CoM dynamics in the x-y plane of a bipedal robot can be approximated using a simple linear inverted pendulum (LIP) model~\cite{kajita20013d}. When the stance ankle is passive, akin to a point contact, the continuous dynamics within a walking step become autonomous, meaning that the progression of the robot’s CoM states depends solely on its state at the beginning of the step. At impact, the CoM position $(p_x, p_y)$ is reset relative to the new stance foot, thereby updating the initial conditions for the subsequent step. Consequently, our MPC planner formulates an optimal control problem for gait planning based on these step-to-step discrete dynamics, using the swing foot stepping positions $(f_x, f_y)$ as control inputs. Additionally, as introduced in our previous work \cite{peng2024real}, the heading angle $\phi$ is incorporated into the state as a single integrator of the turning rate $\omega$.
Let $\mathbf{x} \coloneqq [p_x, v_x, p_y, v_y, \theta]^T\in\mathcal{X}\subset\mathbb{R}^5$ denote the state vector and $\mathbf{u} \coloneqq [f_x, f_y, \omega]^T\in\mathcal{U}\subset\mathbb{R}^3$ represent the control inputs. Assuming a constant step duration $T$, the step-to-step LIP dynamics with heading angle can be expressed as a discrete time linear control system:
\begin{align}
\label{eq:system_dynamics}
    \mathbf{x}_{k+1} =  \mathbf{A_L}
    \mathbf{x}_k
    + \mathbf{B_L}\mathbf{u}_k.
\end{align}
To determine optimal stepping positions and turning rates that enables the robot to follow dynamic subgoals, we formulate an LIP-based Model Predictive Control (LIP-MPC) as the low-level planner. We define a quadratic cost function based on the Euclidean distance between the robot’s CoM position at the beginning of every step and the subgoal $(c_x,c_y)$ as follows:
\begin{align}
\label{eq:cost_func_qp_form}
    q(\mathbf{x}_{k})=&  \left(p_{x_k}-c_{x}\right)^2+\left(p_{y_k}-c_{y}\right)^2 \quad \forall k\in[1,N],
\end{align}
with $N$ being the number of prediction steps. In addition, our formulation incorporates several kinematic and path constraints—including those on velocity, reachability, and maneuverability—to ensure feasible and safe motions. For further details on the construction of these constraints and the overall MPC formulation, please refer to our previous work in \cite{peng2024real}. In this work, we set the robot’s desired height to $H = 1$, the step duration to $T = 0.4$ s, and a prediction horizon to $N = 3$.

Since these constraints are evaluated in the local coordinate frame, they are inherently nonlinear. To linearize them, we map the direction to the subgoal $\phi_c$ to the turning rate control $\omega_k$ at each prediction step in the MPC using the relationship:
\begin{align}
\label{eq:turning_rate_mapping}
    \omega_k=\frac{\phi_c}{NT}.
\end{align}
Benefiting from the high-level RL planner generating dynamic subgoals at each step, the resulting optimization problem in the low-level MPC reduces to a Linear Constraints Quadratic Problem (LCQP). This approach enables the MPC to stabilize the walking process effectively by updating the stepping positions at a frequency of 20 Hz or faster.

\section{Policy Learning with Data Bootstrapping}
\label{sec:setup}

This section describes our reinforcement learning setup and procedure for developing a dynamic subgoal policy in humanoid navigation. To enhance learning efficiency, we employ a data bootstrapping approach that leverages offline expert demonstrations to initialize the learning process.

\begin{figure}
\centering
\vspace{2mm}
    \includegraphics[trim={0cm 0cm 0cm 0cm},clip,width=1\columnwidth]{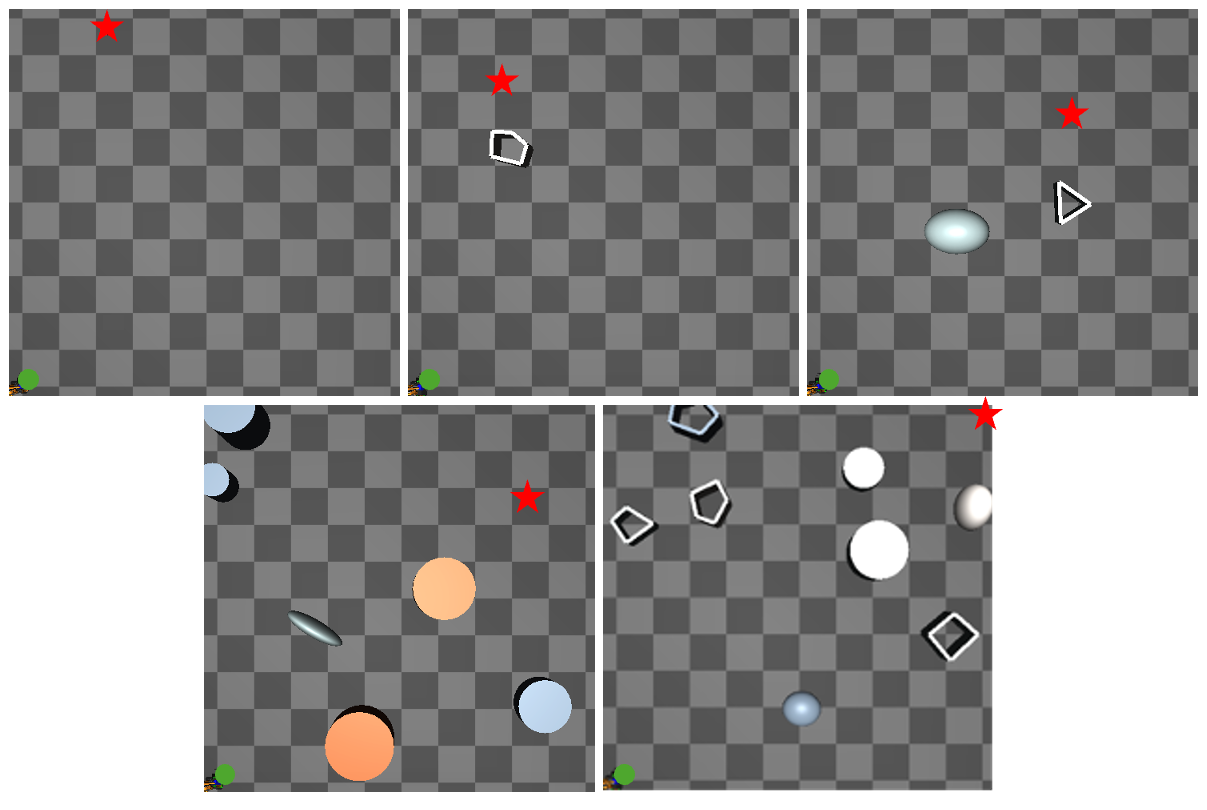}
\caption{Five representative environments from the 50 randomly generated training environments. Here, green dots represent initial positions, and red stars represent goal positions.}
\label{fig:part_env}
\vspace{-2mm}
\end{figure}

\newsec{Training Setup.} We use the Agility Robotics humanoid Digit as our testing platform in a MuJoCo simulation. For training the high-level policy, we generated $50$ randomly generated environments, each starting from the same initial position at \((0,0)\,\text{m}\) and featuring a randomly assigned goal position. Obstacles were placed within a \(10 \times 10\,\text{m}\) area and varied in size, location, and shape (randomly selected from circles, ellipses, or polygons). Among these 50 scenarios, 10 contained no obstacles, while the remaining 40 were evenly distributed across those containing 1, 2, 6, or 8 obstacles (10 scenarios each). \figref{fig:part_env} illustrates five representative settings of these scenarios. \tabref{table:RLparam} presents the hyper-parameters used in the reinforcement learning process, and \tabref{table:rewardparam} shows scaling factors used in the reward design. The learning is performed on an Ubuntu 20.04 laptop with an NVIDIA RTX 4060 GPU.

\begin{table}[ht]
\centering
\begin{tabular}{l l l } \hline
Symbol & meaning & Value \\ \hline
 $\gamma$ & Discount factor & 0.99 \\
 $\alpha_a$ & Learning rate (actor network) & $5e-4$ \\
 $\alpha_c$ & Learning rate (critic network) & $1e-3$ \\
 $N$ & Batch size  & 64\\
 $D$ & Reply buffer size   & 100,000 \\
 $\eta$  & target entropy  & -0.5 \\
\hline
\end{tabular}
\vspace{5mm}
\caption{RL hyper-parameters used in learning}
\label{table:RLparam}
\vspace{-4mm}
\end{table}

\begin{table}[ht]
\centering
\begin{tabular}{l l  } \hline
Reward Parameters & Value \\ \hline
$[a_g, b_g]$ & [2.33, 0.3] \\
$a_\theta$ & 1.39\\
$q_a$ & 0.5 \\
$[q_v, a_v, b_v]$ & $[0.7, 15, 0.5]$\\
$\zeta$ & $0.4$ \\
\hline
\end{tabular}
% \vspace{2mm}
\caption{Reward parameter value}
\label{table:rewardparam}
\vspace{-2mm}
\end{table}

\newsec{Data Bootstrapping.} 
% The LMPC can achieve real-time navigation because its heading angle precomputation mechanism fixes the turning rate for each prediction horizon, linearizing the nonlinear constraints. However, this precomputation mechanism limits navigation flexibility, as it adjusts the turning rate solely toward the goal position. To enhance agility, we introduced the RRT algorithm as a subgoal generator. In contrast, our Carrot-Chasing framework employs the SAC algorithm. 
While the SAC algorithm promotes exploration through its off-policy formulation, its convergence to higher rewards can be hindered if initial explorations are ineffective (e.g., rarely reaching the goal). To address this, we initialize the replay buffer with a demonstration dataset to bootstrap online policy learning. Specifically, we leverage the linear LIP-MPC controller from our previous work \cite{peng2024real}, augmented with a Rapidly-exploring Random Tree (RRT) for enhanced global exploration (denoted as RRT-LMPC). The structure of RRT-LMPC is similar to our proposed hierarchical framework, in which RRT is responsible for generating subgoals for the LMPC. 
Specifically, we collected $10,000$ state-action transactions from the RRT+LMPC controller across various training environments. During the early phase of training, $80\%$ of the replay buffer samples are drawn from this offline demonstration dataset, while the remaining $20\%$ originate from online interactions with the simulation environment. As training progresses, the reliance on the demonstration dataset is gradually reduced until the replay buffer is composed entirely $(100\%)$ of online samples. The overall training process spans 10,000 episodes. Results in the following section demonstrate that the data bootstrapping method significantly improves both learning convergence and navigation performance (e.g., success rate and travel time).
\section{Simulation Results}
\label{sec:sim}
This section details the simulation setup and presents the results that demonstrate the effectiveness and performance of the proposed method in navigating cluttered environments
\footnote{A video showing all simulation results can be found in the \url{https://youtu.be/cbz4vPvsF4g}.}.

% \subsection{Simulation Setup}

\newsec{Simulation Setup.} 
We evaluate the performance and generalization of our approach in two testing environments: a \emph{testing-seen} environment and a \emph{testing-unseen} environment. The \emph{testing-seen} environment comprises all navigation scenarios used during training, while the \emph{testing-unseen} environment is designed to assess the generalization capability of the proposed method. Specifically, the \emph{testing-unseen} environment consists of 25 newly generated scenarios created using the same randomization procedure as the training set; however, each unseen scenario contains eight obstacles (i.e., the maximum difficulty level) of varying shapes and sizes. In all \emph{testing-unseen} cases, the robot starts at \((0,0)\,\text{m}\) and aims to reach the universal goal at \((10,10)\,\text{m}\), which can increase the likelihood of encountering randomly generated obstacles, providing a more rigorous evaluation.

 % \tabref{table:RLparam} \ref{table:rewardparam} \ref{table:MPCparam} presents the selected values of weights and limits used throughout all the tests in this paper. 
 % The simulation runs on Ubuntu 20.04 with an NVIDIA RTX 4060 GPU.

% \begin{table}[h]
% \vspace{2mm}
% \centering
% \begin{tabular}{l l  } \hline
% Low-Level Parameters & Value \\ \hline
% $[v_{x_\mathrm{min}}, v_{x_\mathrm{max}}]$ & $[-0.1, 0.8] $ m/s \\
% $[v_{y_\mathrm{min}}, v_{y_\mathrm{max}}]$ & $[0.1, 0.4]$ m/s\\
% $l_{\mathrm{max}}$ & $0.1\sqrt{3}$m \\
% $\psi$ & $3.6$\\
% \hline
% \end{tabular}
% \caption{Low-level Learning MPC parameters}
% \label{table:MPCparam}
% \vspace{-2mm}
% \end{table}

\subsection{Comparisons against Baseline Model-Based Approaches}
In this section, we evaluate the navigation performance of the proposed RL approach against three baseline model-based methods: a model-based nonlinear LIP-MPC controller~\cite{peng2024unified}, a linear LIP-MPC controller (denoted as LMPC)~\cite{peng2024real}, and the RRT-LMPC---the expert approach used for collecting the demonstration dataset.

\newsec{Real-Time Computation.} The nonlinear LIP-MPC optimizes both stepping positions and turning rates based on the global goal position and obstacle information. However, the inherent nonlinearity of its optimization constraints results in an average computation time of $110\pm57$ ms per optimization, which is insufficient to achieve the necessary real-time update frequency---at least 20 Hz---for stable walking~\cite{castillo_template_2023}. 
In contrast, the linear LIP-MPC (LMPC) efficiently computes gait controls in only $0.94\pm 0.07$ ms, enabling real-time reactive gait generation. Despite this, LMPC may suffer from steering inflexibility due to its precomputation of turning rates~\cite{peng2024real}. To address this limitation, our high-level RL planner---and, alternatively, a global RRT planner---provides dynamic subgoals that guide the low-level LMPC. Notably, the RL planner computes a subgoal from the state and environmental feedback in just $1.3\pm0.5$ ms, which is adequate for our chosen frequency of $2.5$ Hz (i.e., updating at every walking step). We intentionally maintain this lower update frequency, despite the fast computation times, to allow sufficient processing time for environmental and state feedback.

% The nonlinear LIP-MPC controller receives the goal position and obstacle information and constructs an optimization problem with multiple nonlinear constraints based on the robot's state. However, the gait controls, including the turning rate, were computed using nonlinear MPC, requiring an average of $110\pm57$ ms per computation. This computation time is insufficient to support the necessary real-time update frequency—at least 20 Hz—which has been shown to be essential for stable walking in most experiments~\cite{castillo_template_2023}. 
% In contrast, our proposed subgoal pursuit framework significantly improves computational efficiency. The high-level RL planner requires only $1.3\pm0.5$ ms to compute a subgoal, which is updated once per walking cycle. 
% By leveraging RL to handle nonlinearities at the planning level and adopting a hierarchical structure with different update frequencies, our approach effectively reduces the computational burden while maintaining real-time performance.

\begin{figure}
\centering
\vspace{2mm}
    \includegraphics[trim={0cm 0cm 0cm 0cm},clip,width=1\columnwidth]{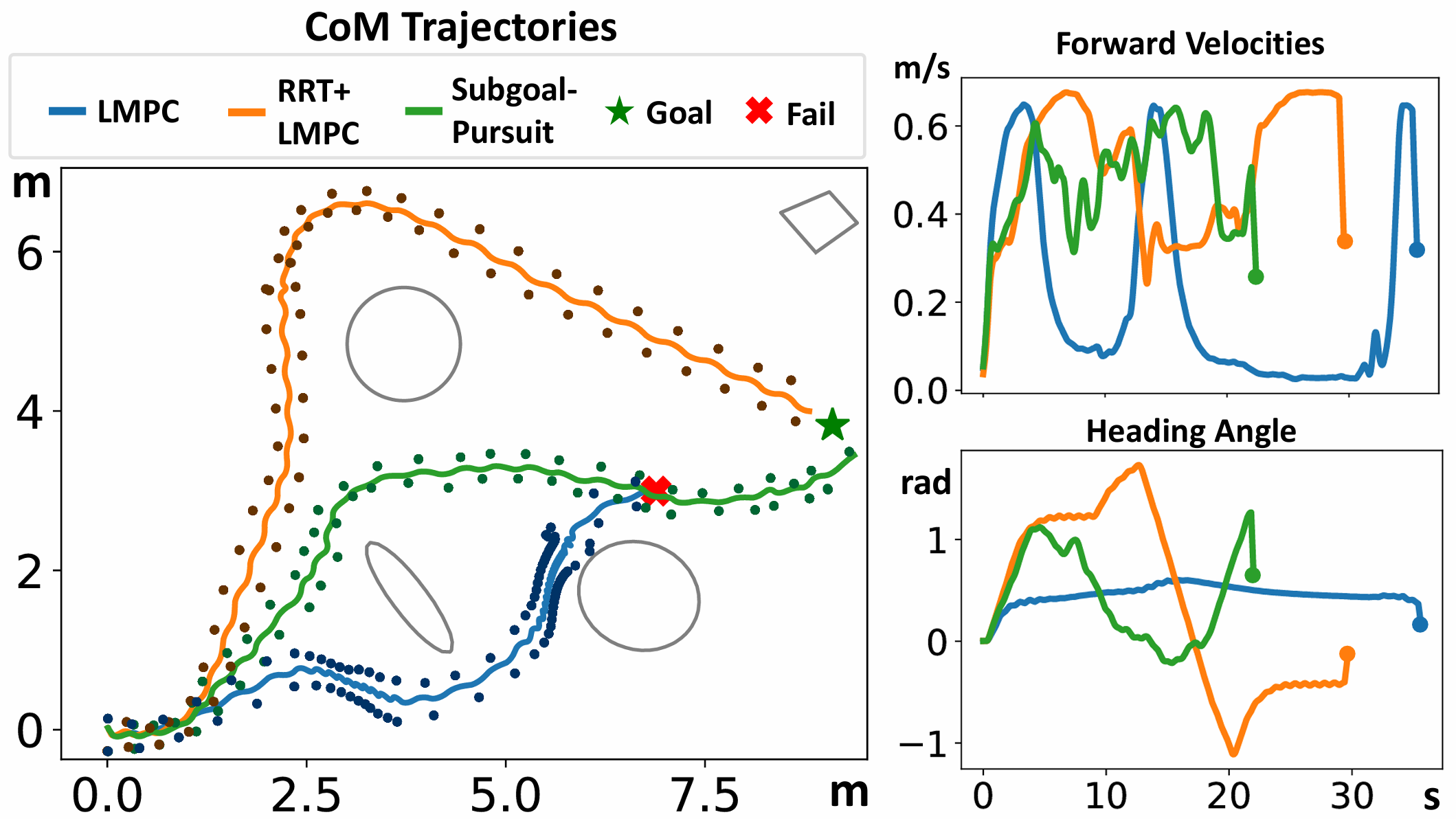}
\caption{The navigation performance of LMPC, RRT-LMPC, and Subgoal Pursuit planner in one specific environment. The right figure shows the resulted paths with stepping positions, and the left plots show the forward walking velocity and commanded turning rates.}
\label{fig:trajectory_comp}
\vspace{-5mm}
\end{figure}

\newsec{Navigation Performance.}
We evaluated the navigation performance of three methods---LMPC without subgoals, RRT+LMPC, and the proposed Subgoal Pursuit planner---across four trials in the \emph{testing-seen} environments. Performance metrics included success rate, accumulated reward, and the average ratio of goal-reaching time (only considering successful trails) relative to our Subgoal Pursuit method. The comparison results are summarized in \tabref{table:structure compare}, with results from one specific trail illustrated in \figref{fig:trajectory_comp}.

\begin{table}[t]
\vspace{3mm}
\centering
\begin{tabular}{l l l l} \hline
Method & \thead{Success\\rate [\%]} & \thead{Accumulated\\Reward} & \thead{Time\\ Ratio}\\ \hline
LMPC &  $84\pm 0.0$ & $91.06\pm 0.0$ & $1.17$\\
RRT+LMPC & $90 \pm 1.4$ & $96.11\pm 3.64$ & $1.02$ \\
\textbf{Our method} & $\mathbf{93.3\pm 2.1}$ & $\mathbf{101.07\pm 2.41}$ & $\mathbf{1}$ \\
\hline
\end{tabular}
\caption{The results of four trials in the \emph{testing-seen} environment for each methods.}
\label{table:structure compare}
\vspace{-2mm}
\end{table}

\begin{figure*}
\centering
% \vspace{2mm}
    \begin{subfigure}[b]{0.32\linewidth}
        %  \centering
         \includegraphics[trim={3.2cm 0cm 3.2cm 0cm},clip,width=1\linewidth]{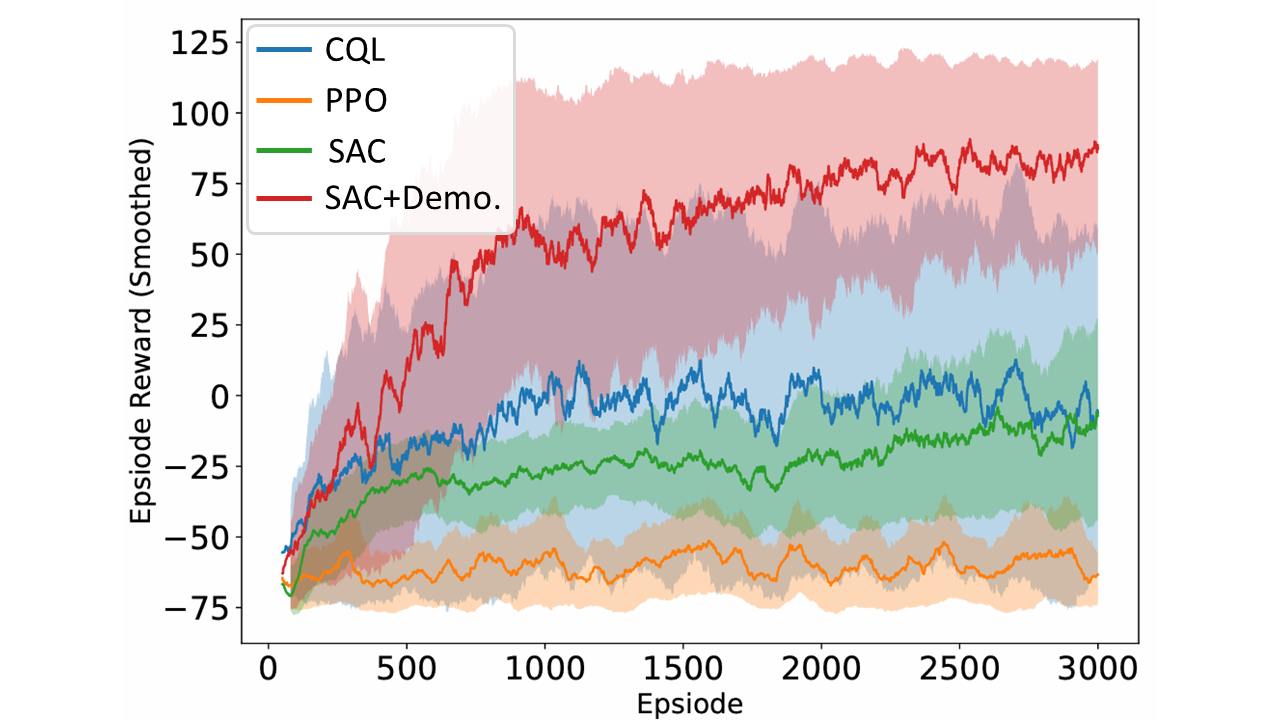}
         \caption{\small{Reward convergence in training}}
         \label{fig:return}
     \end{subfigure}
     \begin{subfigure}[b]{0.32\linewidth}
        %  \centering
        \includegraphics[trim={3.1cm 0cm 3.1cm 0cm},clip,width=1\linewidth]{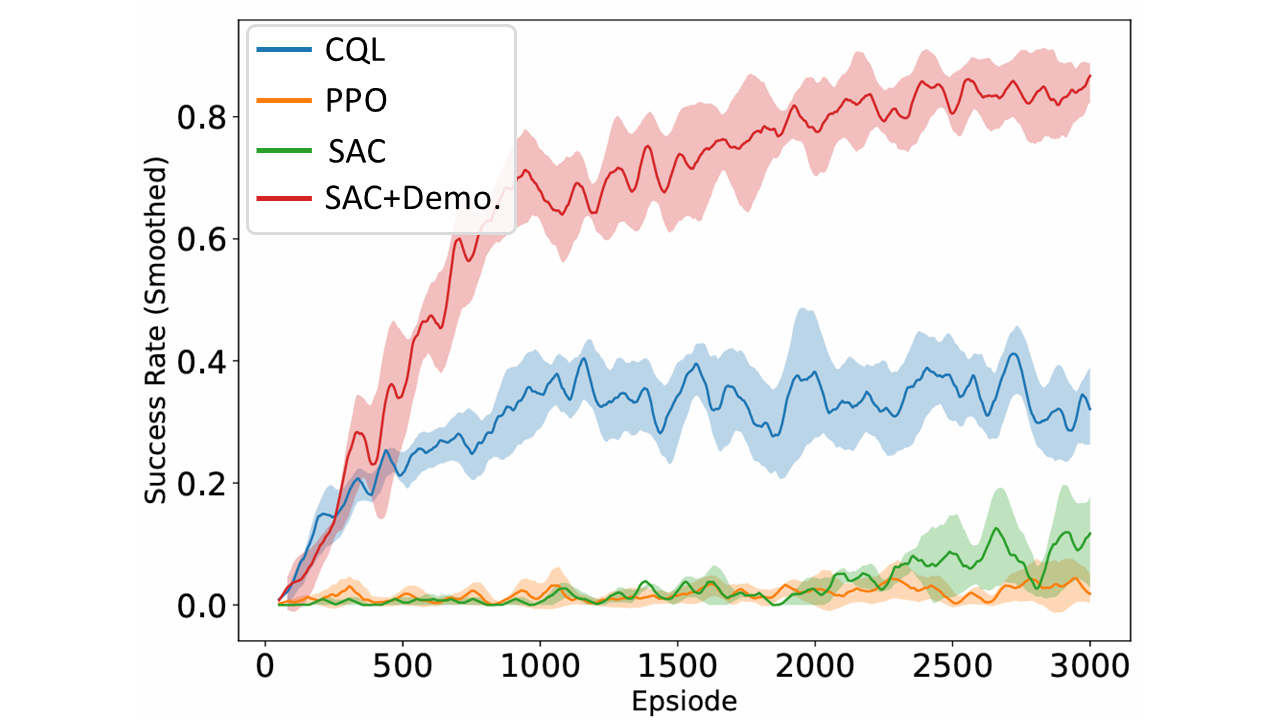}
         \caption{\small{Success rate of reaching goals in training}}
         \label{fig:succ}
     \end{subfigure}
     \begin{subfigure}[b]{0.32\linewidth}
        %  \centering
        \includegraphics[trim={3.8cm 0cm 4cm 0cm},clip,width=0.94\linewidth]{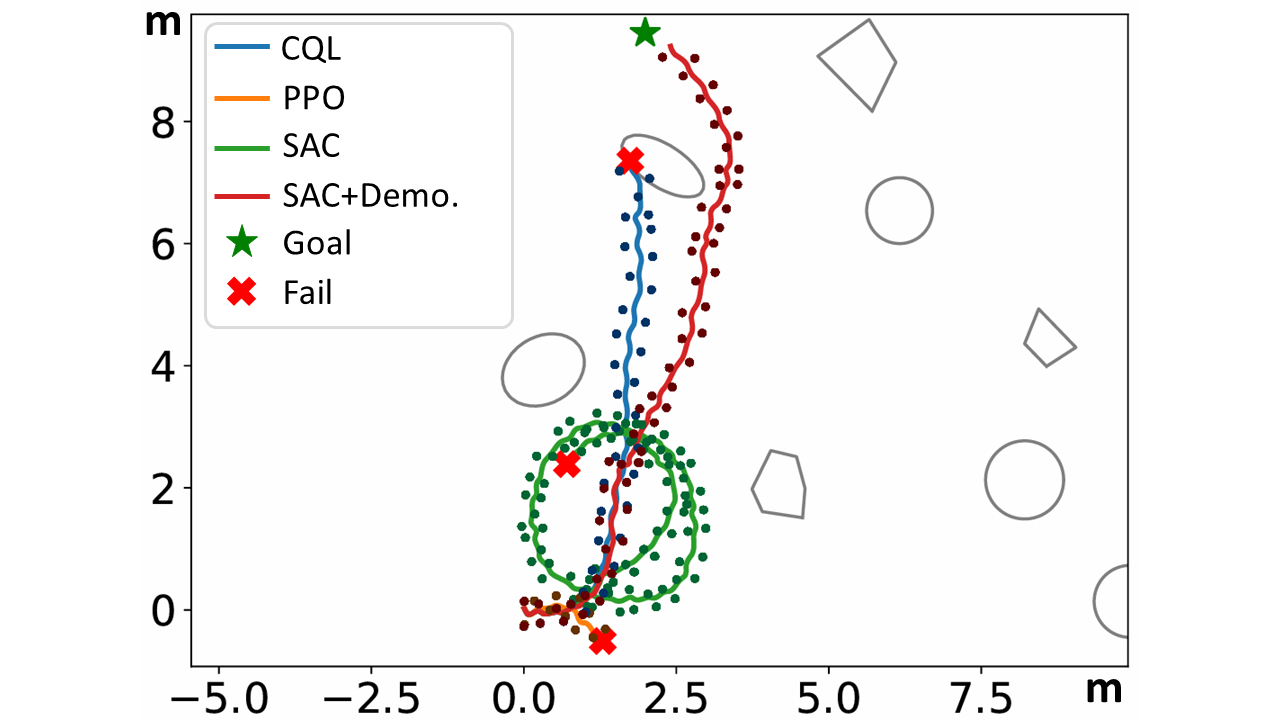}
         \caption{\small{The executed paths in testing scenario}}
         \label{fig:traj}
     \end{subfigure}
\caption{{Comparing training performance of four different learning-based approaches.}}
\label{fig:baseline_comp}
\vspace{-4mm}
\end{figure*}

From \tabref{table:structure compare}, we observe that the linear MPC approach, being the most conservative, achieves the lowest success rate. Its limited ability to perform agile maneuvers often requires the robot to slow down or rely on side-walking to avoid obstacles, which increases the likelihood of time-outs or becoming trapped in local minima. The introduction of a global planner (e.g., RRT) improves the flexibility and success rate, as evidenced by higher cumulative rewards and average speed. However, the inherent randomness of RRT and its lack of consideration for the bipedal robot's dynamic model often lead to the generation of overly aggressive subgoals, thereby limiting further improvements in success rate. 
In contrast, our proposed Subgoal Pursuit method achieves the highest success rate and accumulated reward among the three approaches, and it also reaches the goal fastest on average compared to the other two methods.
\figref{fig:trajectory_comp} shows the navigation trajectories of the three approaches in a specific trail. In this case, LMPC fails due to a timeout caused by slow lateral walking. RRT+LMPC reaches the goal but follows the longest trajectory. 
% Due to the randomness of the RRT algorithm, it requires more aggressive actions and a longer arrival time. 
Although RRT+LMPC maintains the highest forward velocity when moving straight, its efficiency is compromised by unnecessary detours. In contrast, the Subgoal Pursuit method achieves the goal the fastest by more smoothly regulating both the walking speed and turning rate. 
% However, this flexibility comes at the cost of slightly reduced straight-line speed.

By leveraging RL as a high-level planner, our algorithm determines robust subgoals through continuous exploration in the environment. Notably, bootstrapping the RL training with a model-based dataset results in final performance surpassing that of the original model-based approach, underscoring the strength of RL in navigation tasks. It is important to note that our approach relies solely on local map data, in contrast to the global map used by RRT. This reliance on local information introduces a slightly higher variance, which may also stem from the inherent randomness and uncertainty of the policy structure.
% However, while Carrot-Chasing achieves a higher average success rate, it also exhibits greater variance, indicating some level of randomness and uncertainty in the learned policy. 

% Additionally, although the first two model-based algorithms can maintain a faster speed when going straight, they are limited by flexibility constraints or algorithm randomness, and the average time they spend to reach the goal is greater than our method.

\subsection{Comparison against Other Learning Based Methods}
To identify the most effective RL algorithm for the high-level planning component of our Subgoal Pursuit framework, we evaluated several state-of-the-art learning-based methods. Each method was trained and tested under consistent conditions and compared based on key performance metrics such as success rate, accumulated reward, and average reaching time. The following sections provide detailed descriptions of these baseline algorithms and discuss their relative performance within our framework.

\newsec{Conservative Q-Learning(CQL):} CQL \cite{kumar2020conservative} is an offline RL algorithm that learns from pre-collected data by conservatively estimating Q-values, which helps ensure stability and reliability without the need for online interaction.

\newsec{Proximal Policy Optimization (PPO):} PPO \cite{schulman2017proximal} is an on-policy algorithm that uses a clipped objective function to constrain policy updates, ensuring stable training dynamics.

\newsec{Soft Actor-Critic (SAC):} SAC is an off-policy actor-critic algorithm that optimizes a stochastic policy with an entropy-augmented reward. In our tests, we use SAC as a baseline to showcase the benefits of bootstrapping online RL with prior demonstration data.

% \newsec{Hindsight Experience Replay(HER)} is a technique that improves sample efficiency in sparse-reward environments by relabeling failed experiences as successful ones, accelerating learning.
% \newsec{Generative Adversarial Imitation Learning(GAIL):} GAIL is one of the SOTA BC method. By leveraging an adversarial framework, GAIL effectively mimics expert behavior, offering a high-performance benchmark for imitation learning approaches. 

% The training reward of each learning based approaches are shown in \figref{fig:baseline_comp}. In the first plot, we We compared the results of three RL algorithms using the same dataset. We found that SAC as an off-policy RL algorithm is more suitable for robot navigation environments than A2C and PPO. In the second plot, we compared the results of the SAC algorithm using different techniques. We found that using HER or dataset bootstrapping greatly improved the learning efficiency. Although HER has outstanding performance as a SOTA technique in goal-oriented RL, using dataset bootstrapping will provide better training results when there is a dataset. The last plot shows the training results of online RL, offline RL and Behavior Cloning methods. We found that the SAC+data bootstrap method still performed best, which shows that the interaction between the learning-based algorithm and the environment is very important for improving the performance of the policy.

\figref{fig:baseline_comp} shows the comparison of training outcomes of the four learning-based approaches: CQL, PPO, SAC, and SAC with demonstration. 
PPO consistently underperforms due to its on-policy nature limits sample efficiency in complex tasks. SAC benefits from its off-policy formulation, allowing it to reuse transitions more effectively. 
However, SAC’s training performance is highly sensitive to the quality of collected transitions. 
In our experiments, when the agent rarely reached the goal during exploration, SAC tended to wander in circles until timing out, as illustrated in \figref{fig:traj}. Conversely, when the goal was reached more frequently, high-quality transitions stored in the replay buffer helped guide SAC toward a better policy. Consequently, SAC’s episodic reward often converged to levels comparable to CQL, while its average success rate remained relatively low.

CQL, benefiting from its offline RL formulation and efficient use of demonstration data, achieved faster improvements in both episodic reward and success rate during the early training stages. However, due to its lack of interaction with the environment during training, CQL struggled to adapt to dynamic environmental changes, leading to occasional obstacle collisions, as shown in \figref{fig:traj}. 
SAC with demonstration outperformed all other baselines by combining demonstration bootstrapping and interactive exploration. Leveraging offline demonstrations directed the agent toward more promising strategies early in training, thereby reducing the reliance on random exploration and avoiding suboptimal behaviors. This complementary effect accelerated learning and stabilized the training process, ultimately resulting in improved performance.

To further evaluate the performance and generalization of each learning-based approach, we tested the trained policies both in \emph{testing-seen} environments and \emph{testing-unseen} environments.  The results are summarized in \tabref{table:RL50case} and \tabref{table:RL25case}. We observe that SAC with demonstration achieves the highest success rate and accumulated reward in both testing, demonstrating superior performance and generalization over other baselines. 
PPO has difficulty learning navigation strategies in both cases due to its low sample efficiency. SAC has a slightly better performance than PPO in these tasks but is still significantly worse than SAC with demonstration. While CQL has lower success rates and accumulated rewards than our proposed approach in both testing, it results in the fastest time to reach the goal when successful. 

% To evaluate generalization to unseen scenarios, we tested the learning-based high-level planners in \emph{testing-unseen} environment. The results, summarized in \tabref{table:RL25case}, show that the success rates of all learning-based methods decline in this testing, but SAC+demonstration still maintains a high success rate,  indicating strong generalization ability in navigating previously unseen environments.

\begin{table}[h]
\vspace{3mm}
\centering
\begin{tabular}{l l l l} \hline
Method & \thead{Success\\rate [\%]} & \thead{Accumulated\\Reward} & \thead{Time\\Ratio}\\ \hline
CQL & $44.3\pm 4.2$ & $14.32\pm 5.97$ & $\mathbf{0.91}$ \\
PPO &  $0\pm 0$ & $-73.43\pm 4.64$ & $-$\\
SAC & $9 \pm 2.6$ & $-32.72\pm 3.57$ & $1.01$ \\
\textbf{SAC+Demo.} & $\mathbf{93.3\pm 2.1}$ & $\mathbf{101.07\pm 2.41}$ & $1$ \\
\hline
\end{tabular}
\caption{The results of four trials in the \emph{testing-seen} environment for each method.}
\label{table:RL50case}
\vspace{-3.5mm}
\end{table}

\begin{table}[h]
\vspace{3mm}
\centering
\begin{tabular}{l l l l} \hline
Method & \thead{Success\\rate [\%]} & \thead{Accumulated\\Reward} & \thead{Time\\Ratio}\\ \hline
CQL & $14\pm 8.2$ & $-48\pm 3.53$ & $\mathbf{0.92}$ \\
PPO &  $0\pm 0$ & $-75.43\pm 1.04$ & $-$\\
SAC & $5 \pm 2$ & $-49.21\pm 2.09$ & $1.08$ \\
\textbf{SAC+Demo.} &  $\mathbf{89.3\pm 2.2}$ & $\mathbf{101.47\pm 4.38}$ & $1$\\
\hline
\end{tabular}
\caption{The results of four trials in the \emph{testing-unseen} environment for each method.}
\label{table:RL25case}
\vspace{-3.5mm}
\end{table}

\section{Conclusion} 
\label{sec:conclusion}

In this paper, we introduced the Subgoal-Pursuit method---a hierarchical navigation framework for bipedal robots that uses RL to learn a high-level planner to manage navigation complexity and used a model-based gait planner to ensure stable locomotion. In addition, we accelerated the training of the SAC algorithm by using a dataset bootstrapping technique. The results demonstrate its effectiveness in guiding bipedal robots through complex environments. However, there remains room for improving trajectory optimization and enhancing robustness and generalization to unseen and dynamic environments. Future work will focus on enhancing the learning framework to guarantee safe navigation in unseen and dynamic environments, as well as implementing the proposed approach on hardware for real-world validation.

% an ESC planner based on the LIP model-based MPC for bipedal robots. This planner possesses capabilities for both path planning and control planning. Building upon this planner, we designed a hierarchical controller with the ability to control the walking process of a bipedal robot, enabling it to navigate and avoid obstacles. However, our algorithm also faces challenges such as the time-consuming resolution of nonlinear kinematic constraints and the discrepancies between simplified model and real hardware. For future work, we plan to enhance the efficiency of our algorithm by restructuring it to reduce the optimization problem-solving time. Moreover, to address potential model mismatch issues, we plan to experiment with new motion models or use learning-based methods to unlock the locomotion potential of bipedal robots.

% \newpage
\bibliographystyle{IEEETran}
\bibliography{references.bib}

\end{document}